\documentclass{article}

     \usepackage[preprint]{neurips_2019}

\usepackage[utf8]{inputenc} 
\usepackage[T1]{fontenc}    
\usepackage{hyperref}       
\usepackage{url}            
\usepackage{booktabs}       
\usepackage{amsfonts}       
\usepackage{nicefrac}       
\usepackage{microtype}      
\usepackage{subcaption}
\usepackage{graphicx}
\usepackage{amsmath}
\usepackage{todonotes}

\title{A COLD Approach to Generating Optimal Samples}

\author{%
  Omar Mahmood \\
  Center for Data Science\\
  New York University\\
  New York, NY 10011 \\
  \texttt{onm217@nyu.edu} \\
  \And
  José Miguel Hernández-Lobato \\
  Department of Engineering \\
  University of Cambridge \\
  Cambridge, UK CB2 1PZ \\
  \texttt{jmh233@cam.ac.uk}
}

\begin{document}

\maketitle

\begin{abstract}
Optimising discrete data for a desired characteristic using gradient-based methods involves projecting the data into a continuous latent space and carrying out optimisation in this space. Carrying out global optimisation is difficult as optimisers are likely to follow gradients into regions of the latent space that the model has not been exposed to during training; samples generated from these regions are likely to be too dissimilar to the training data to be useful. We propose Constrained Optimisation with Latent Distributions (COLD), a constrained global optimisation procedure to find samples with high values of a desired property that are similar to yet distinct from the training data. We find that on MNIST, our procedure yields optima for each of three different objectives, and that enforcing tighter constraints improves the quality and increases the diversity of the generated images. On the ChEMBL molecular dataset, our method generates a diverse set of new molecules with drug-likeness scores similar to those of the highest-scoring molecules in the training data. We also demonstrate a computationally efficient way to approximate the constraint when evaluating it exactly is computationally expensive.
\end{abstract}

\section{Introduction}

Discrete data cannot be directly optimised for a desired quanitifiable property using gradient-based techniques so other, less effective methods are often used. For example, a goal in the drug discovery process may be to produce a molecule with as high of a particular quantifiable chemical property as possible \cite{lobato}. As molecules are discrete graphs, de novo drug design techniques use optimisation techniques such as evolutionary search, simulated annealing and metropolis search \cite{de_novo} to sample suitable molecules from a search space that is on the order of $10^{60}$ molecules \cite{search_space_size}. Following the work of G\'omez-Bombarelli et al \cite{lobato}, discrete data can be mapped to representations in the continuous latent space of a Variational Autoencoder (VAE) \cite{aevb}, following which optimisation can be carried out in this space to yield representations with high values of the desired quantifiable property. These representations can be converted into high-scoring samples.

However, only local optimisation has previously been carried out using the latent space representations of training data points as a starting point \cite{lobato}. These are the areas on which the model has been trained and so they are more likely than other areas to correspond to valid molecules with scores similar to those predicted by the model. Global optimisation cannot be carried out in random regions of the latent space as the optimisation process is likely to end up in untrained regions of the latent space that are unlikely to produce reasonable, high-scoring samples \cite{griffiths}. Optimisation is hence limited to finding improvements on existing training data points rather than generating entirely new, high-scoring samples. In this work we propose Constrained Optimisation with Latent Distributions (COLD), a technique for global optimisation that involves finding trained regions of the latent space and carrying out constrained local optimisation using high-scoring points in these regions as starting points. This should allow for the generation of high-scoring samples that share the characteristics of the training data but are each significantly distinct from any of the individual data points in the training set.

\section{Method}

We jointly train a VAE \cite{aevb} with a predictor as in \cite{lobato}. Throughout the rest of this paper, we refer to this joint model as a Predictive Variational Autoencoder (PVAE). We optimise a lower bound on the log likelihood (ELBO) together with a squared error loss between the true value $w$ and predicted value $f(z)$ of the target property of the data $x$ with latent representation $z$:

\begin{equation}
    Loss(x, w) = -(\mathbb{E}_{z \sim q(z|x)}\log p(x|z) - KL(q(z|x)||p(z))) + (w - f(z))^2
\end{equation}

Once the model is trained, we pass each of the training data points through the encoder to get the mean and variance vectors of a Gaussian distribution. We take each of these distributions as a component of a uniformly weighted Gaussian mixture model. The resulting GMM represents the encoding distribution of the training data. Regions of the latent space where this GMM has a higher density are more likely to have been seen by the model during training than regions where this GMM has a lower density. Therefore points from higher density regions are more likely to produce samples similar to the training data when passed through the trained decoder than are points from low density regions. 

Ideally we would like to evaluate this GMM on a grid of points in the latent space and filter out low density regions. However, the number of parameters in the GMM is $2 \cdot D \cdot N$ where $D$ is the dimensionality of the latent space and $N$ is the number of training data points, and the number of points in the proposed latent space grid scales exponentially with $D$. It is infeasible to hold a grid of sufficient size in memory and to evaluate the GMM density at each point. As a result, we instead sample grid points from different parts of the latent space.

After obtaining grid points with their associated GMM densities, we select a density threshold $\eta$ below which the points with which the densities are associated are unlikely to yield reasonable samples when passed through the decoder. We retain only those points at which the density is higher than the threshold. These points are candidates for local optimisation.

The candidate points are passed through the trained predictor to yield a predicted score for each point. We select the points with the $t$ highest predicted scores and carry out local optimisation using them as initial coordinates for the optimiser. The output point from each local optimisation is passed into the decoder to yield the estimated globally optimum sample. The true score is obtained by calculating the target property of the generated sample in the same way as is done for the training data. The sample with the highest true score is considered as our global optimum. We can improve performance at the cost of computational intensity and memory usage by sampling a large number of points from the latent space and choosing a high value for $t$ during optimisation.

\textbf{K-Nearest Neighbours Density Approximation}

For high dimensional latent spaces with large amounts of training data, it can be computationally expensive to compute the GMM density at each optimisation step. In order to reduce computational burden, an approximate method of evaluating the GMM density can be used. One such method that we employ in certain experiments consists of constructing a k-nearest neighbour graph of the encoded means of the training data in the latent space. This is done using a hierarchical navigable small world network (HNSW) algorithm \cite{nsw}. Once the graph is constructed, the density at an arbitrary point in the latent space is calculated by querying the graph for the $k$ nearest neighbours (where $k$ is determined experimentally) and calculating the density using these neighbours. The contribution of points that are further away is assumed to be zero.

In order to avoid numerical underflow, the log-sum-exp trick is used. In the following equations, $p_i(g)$ is the probability density of the latent space encoding of datapoint $i$ at latent space point $g$, $D$ is the dimensionality of the latent space, $\mu_{id}$ and $\sigma_{id}$ are the $d$th dimension of the encoding mean and variance respectively, $o$ is an overflow constant, and $\hat{p}(g, k)$ is the GMM density at $g$ calculated using the $k$ nearest neighbours:

\begin{align}
&\log(p_i(g)) = -\frac{1}{2\sigma_{id}^2}(g_d - \mu_{id})^2 - \frac{d}{2}\log(2\pi) - \frac{1}{2}\sum_{d=1}^D \log(\sigma_{id}^2) \,, \\
&o = \max_i (\log(p_i(g))) \,, \\
&\log(\hat{p}(g, k)) = \log \left( \sum_{i=1}^{n}\exp(\log(p_i(g)) - o) \right) + o - \log(N) .
\end{align}

The equation is exactly equal to the full GMM density for $k=N$, and approximate for $k<N$.

\section{Experiments and Discussion}

\subsection{MNIST}
\subsubsection{Motivation}
For our first set of experiments, we focus on the MNIST dataset \cite{mnist}. Handwritten digit images are easy to visualise and understand. Furthermore, the dataset is relatively small, allowing us to exactly evaluate the GMM density quickly and easily. As a result, these experiments should be useful in evaluating COLD while gaining a better understanding of how it works.
\subsubsection{Experimental Setup}
We train PVAEs on all images labelled as $3$ in the MNIST dataset. We train three such models with a $N(z; 0, \mathbb{I})$ prior and a  different predictor objective for each. The objectives we use are the thickness, rotation and aspect ratio of each image. We calculate these objectives as follows:

\textbf{Thickness:} We use the average value of the image's pixel intensities as a measure of its thickness.

\textbf{Aspect Ratio:} We define the ratio $\frac{width}{height}$ of the image as its aspect ratio, with $m$ as the maximum possible pixel intensity, and $width$ and $height$ defined as follows:

\begin{equation}
    \begin{split}
    &height(x) = max \{i: ||x_i||_\infty > \frac{m}{2}\} - min \{i: ||x_i||_\infty > \frac{m}{2}\} \\
    &width(x) = max \{j: ||(x^T)_j||_\infty > \frac{m}{2}\} - min \{j: ||(x^T)_j||_\infty > \frac{m}{2}\} \\
    \end{split}
\end{equation}.

\textbf{Rotation:} We binarise the image by setting pixel values greater than $\frac{m}{2}$ to 1, and setting the rest to 0. We carry out PCA on the binarised image to obtain the second principal component $v_2 = \begin{bmatrix} v_{2_1} \\ v_{2_2}  \end{bmatrix}$. We use the slope $\frac{v_{2_2}}{v_{2_1}}$ of the second principal component as a measure of the image's rotation. The greater the value of the slope, the greater the anticlockwise rotation of the image.

In order to determine the relationship between threshold value and quality of generated images, we define a metric by which to evaluate image quality. We train a digit classifier on the AffNIST dataset \cite{affnist}, which contains images from the original MNIST dataset that have been subjected to affine transformations. We pass images generated by the optimisation process through the classifier and use as our quality metric the confidence of the classifier that the generated image is a $3$. We refer to this as the 3-confidence of the image. High quality images are expected to be predicted as a 3 with high confidence, since they should look similar to a 3. We use the AffNIST dataset to train our classifier as we expect the optimisation process to output images that are transformations of those in the original MNIST dataset, which may not be accurately classified by a classifier trained on MNIST. Optimisation is carried out using the derivative-free COBYLA algorithm \cite{powell_1994}.

\subsubsection{Results and Analysis}

\begin{figure}[h]
    \begin{subfigure}{0.33\textwidth}
        \includegraphics[width=\textwidth]{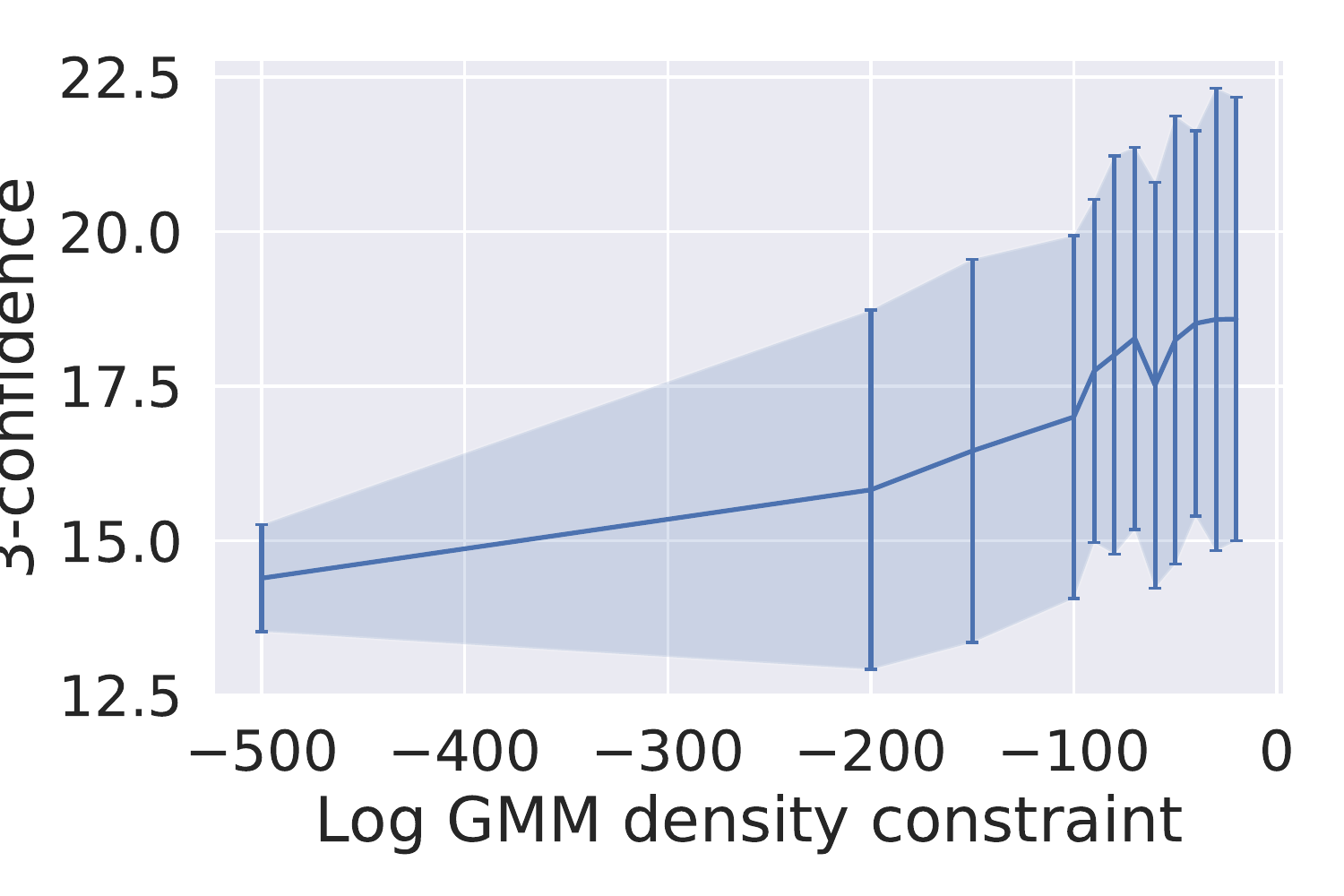}
        \caption{Rotation}
        \label{fig:conf_pca}
    \end{subfigure}
    \begin{subfigure}{0.33\textwidth}
        \includegraphics[width=\textwidth]{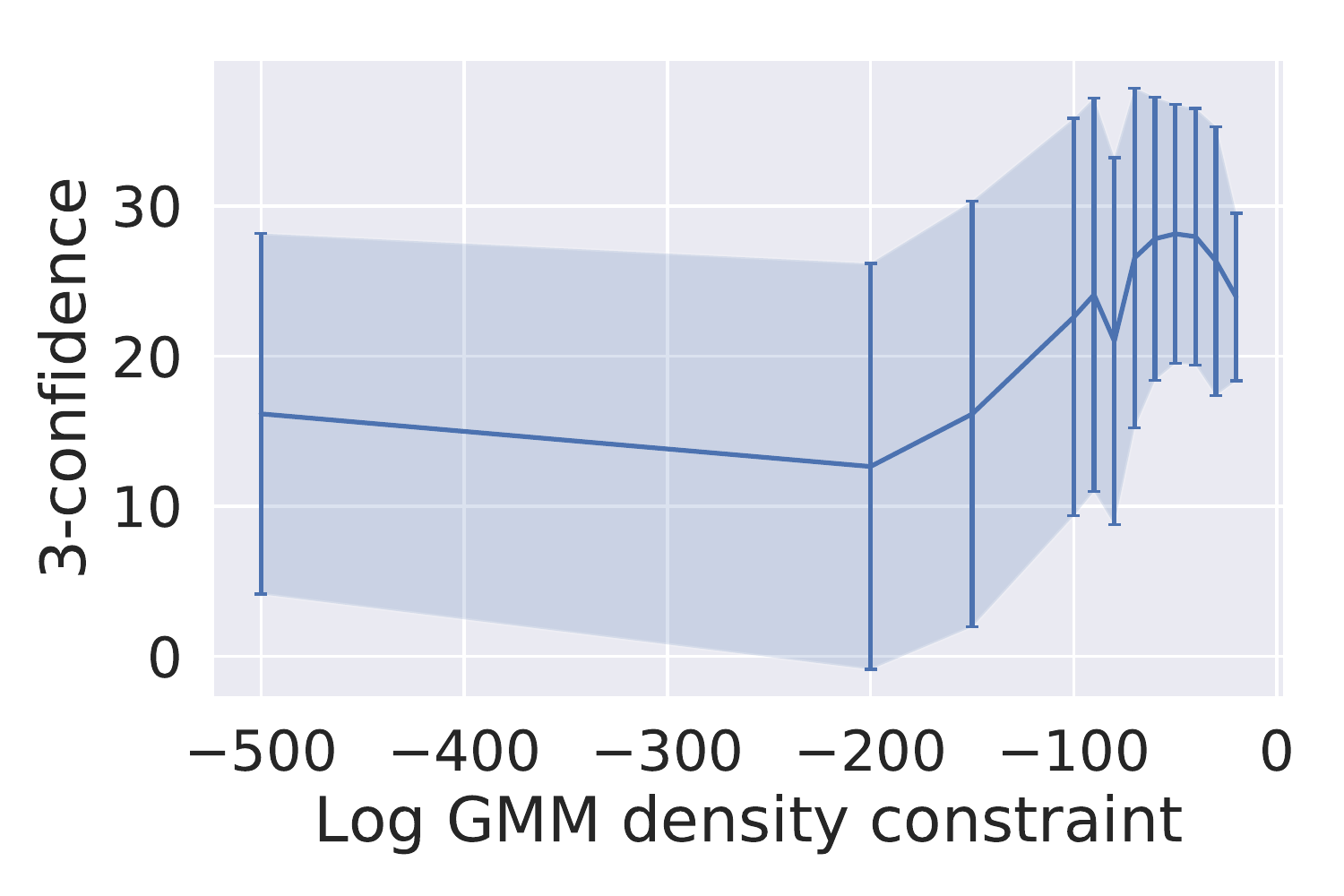}
        \caption{Thickness}
        \label{fig:conf_thickness}
    \end{subfigure}
    \begin{subfigure}{0.33\textwidth}
        \includegraphics[width=\textwidth]{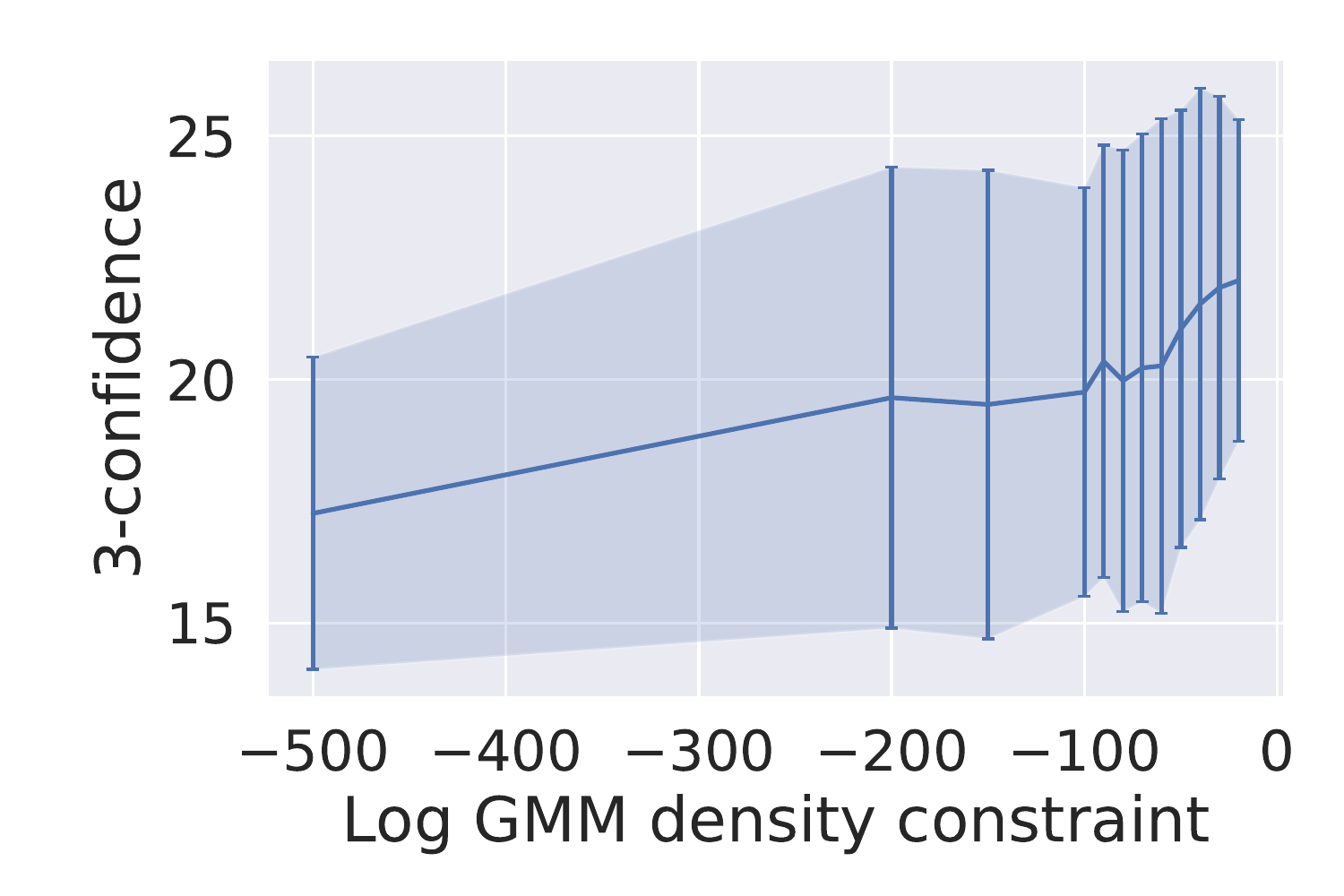}
        \caption{Aspect ratio}
        \label{fig:conf_aspect}
    \end{subfigure}
    \caption{Image quality as a function of log GMM density constraint}
    \label{fig:conf}
\end{figure}

\begin{figure}[h]
    \begin{subfigure}{0.33\textwidth}
        \includegraphics[width=\textwidth]{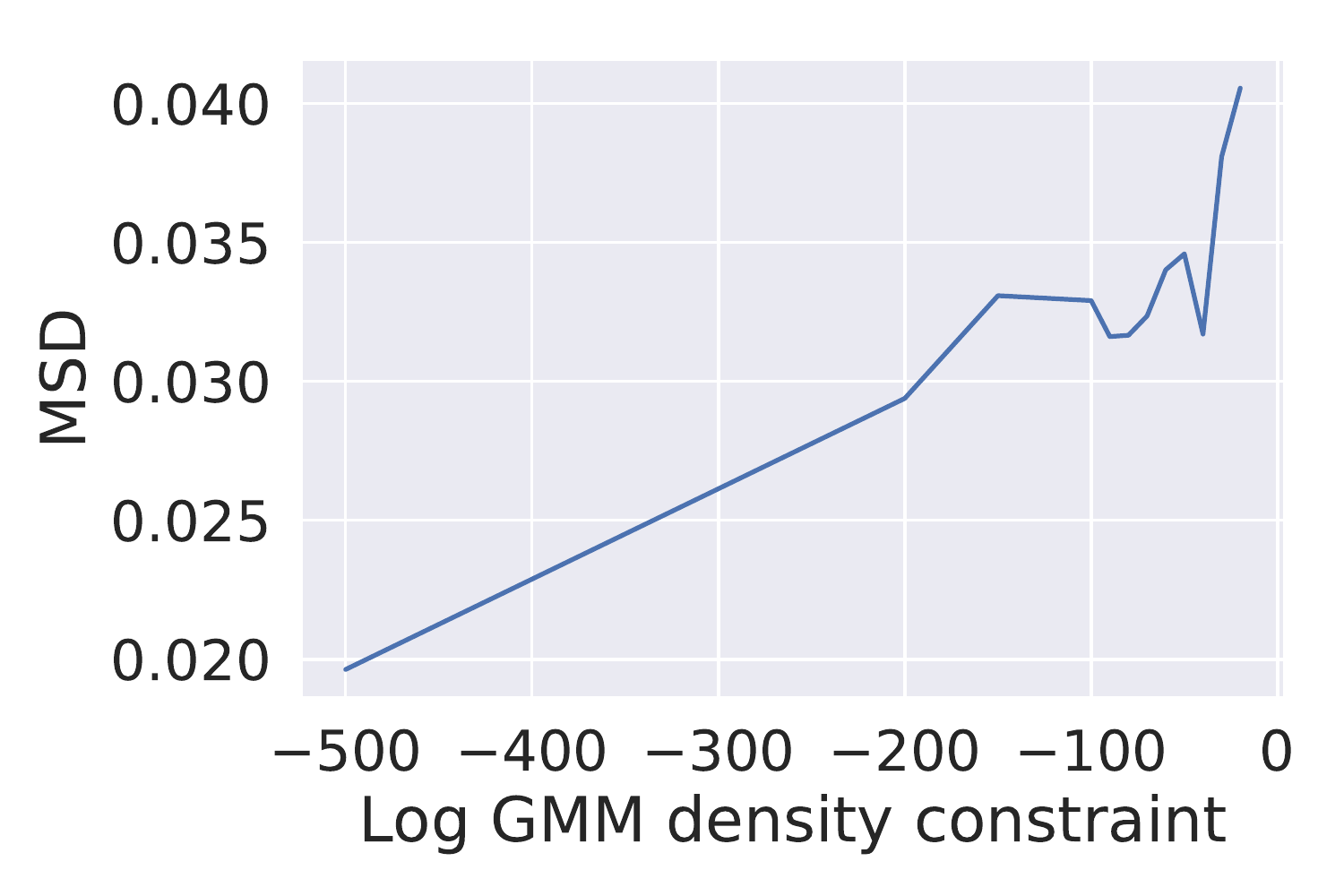}
        \caption{Rotation}
        \label{fig:diversity_pca}
    \end{subfigure}
    \begin{subfigure}{0.33\textwidth}
        \includegraphics[width=\textwidth]{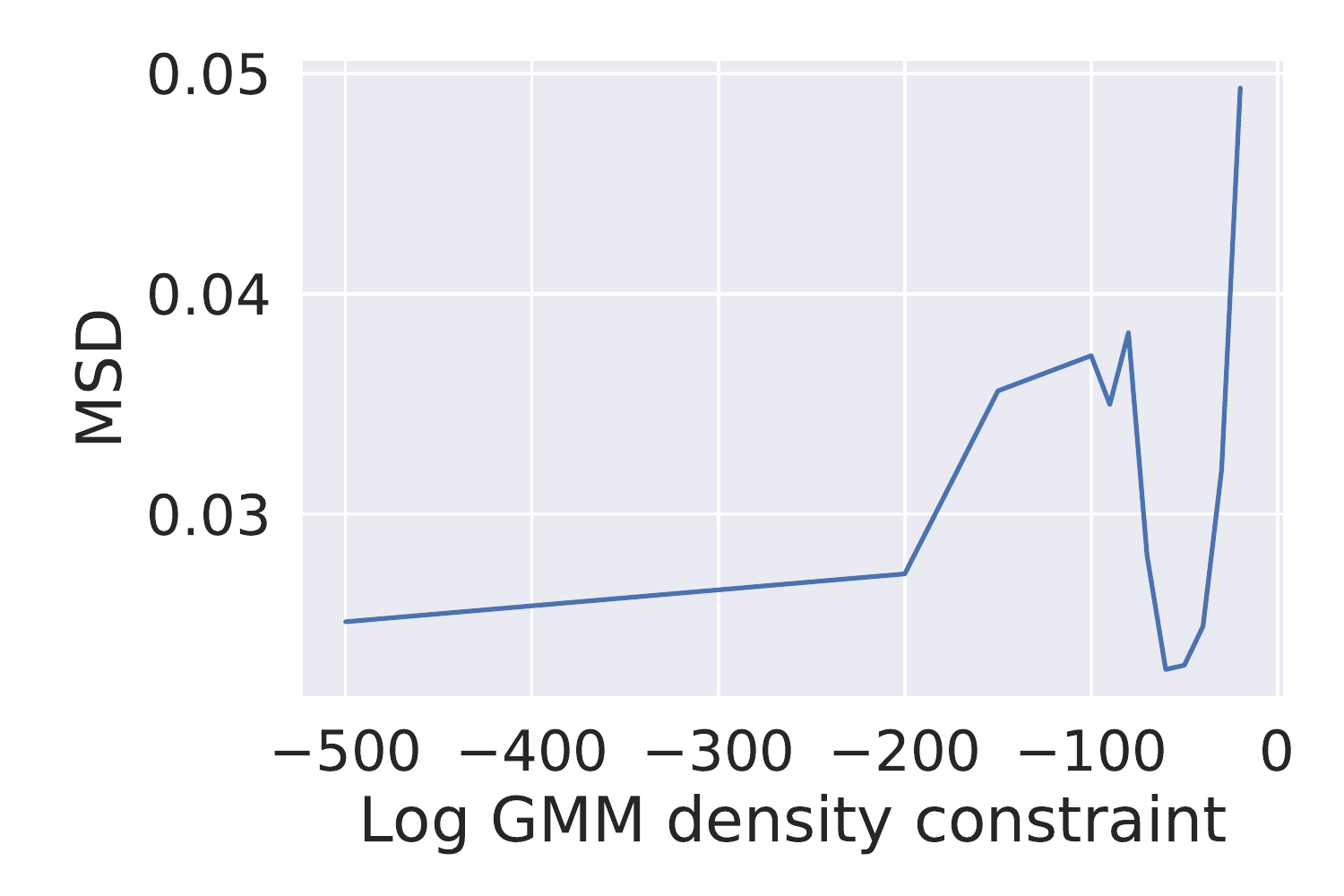}
        \caption{Thickness}
        \label{fig:diversity_thickness}
    \end{subfigure}
    \begin{subfigure}{0.33\textwidth}
        \includegraphics[width=\textwidth]{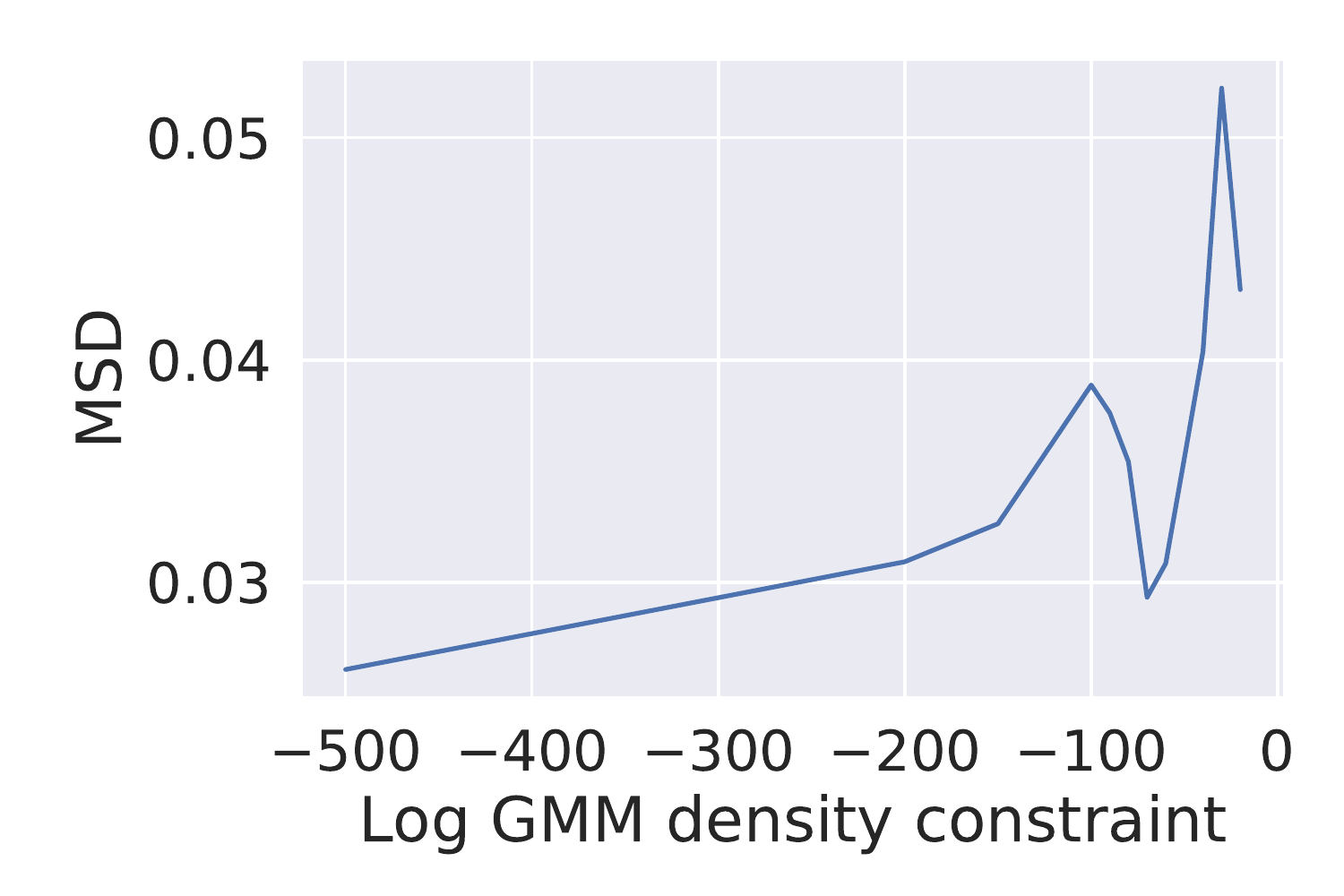}
        \caption{Aspect ratio}
        \label{fig:diversity_aspect}
    \end{subfigure}
    \caption{Image diversity as a function of log GMM density constraint}
    \label{fig:diversity}
\end{figure}

Figure \ref{fig:conf} shows that for all three objectives, the 3-confidence of the optimal images increases as the log GMM density constraint is increased. We discover that as we loosen the density constraint, the optimisation process tends to converge to the same local optimum, and hence the same generated image, for several different initialisation points in the latent space. We evaluate the diversity of the images generated from the optimisation process at different thresholds. We define the diversity $u \in \mathbb{R}$ as the average mean squared distance between images:

\begin{equation}
    u = \frac{2}{t(t-1)}\sum_{i=1}^t \sum_{j=i+1}^t \frac{1}{h^2} ||x_i - x_j||_\mathcal{F}^2
\end{equation}

where each image $x$ is of dimensionality $h \times h$. Figure \ref{fig:diversity} shows the diversity at various thresholds for all three objectives. We observe that the diversity generally increases as the threshold is tightened. If the optimisation is run without constraints, convergence is not achieved in any of these experiments.

\begin{figure}[h]
    \begin{subfigure}{0.33\textwidth}
        \includegraphics[width=\textwidth]{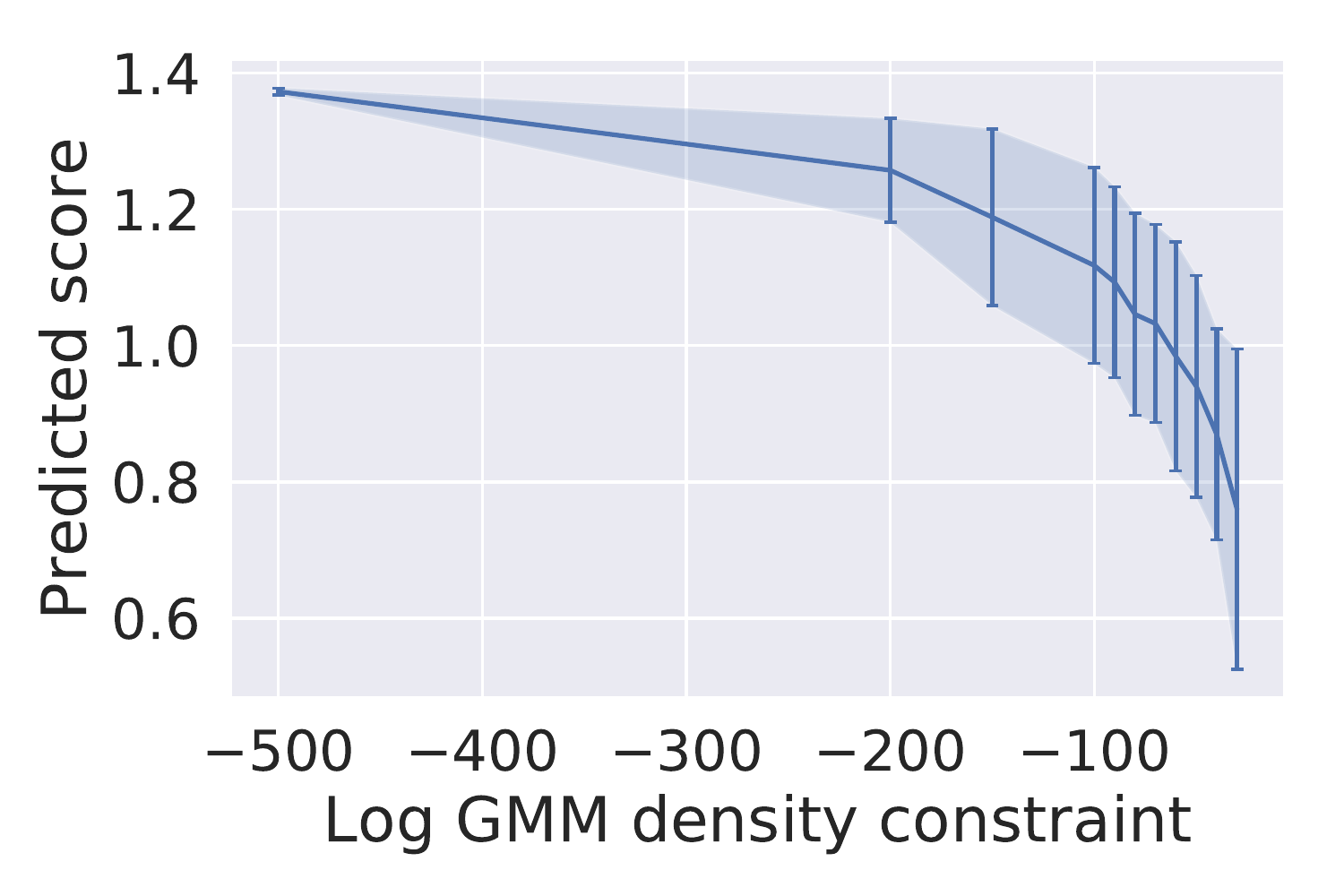}
        \caption{Rotation}
        \label{fig:predscores_pca}
    \end{subfigure}
    \begin{subfigure}{0.33\textwidth}
        \includegraphics[width=\textwidth]{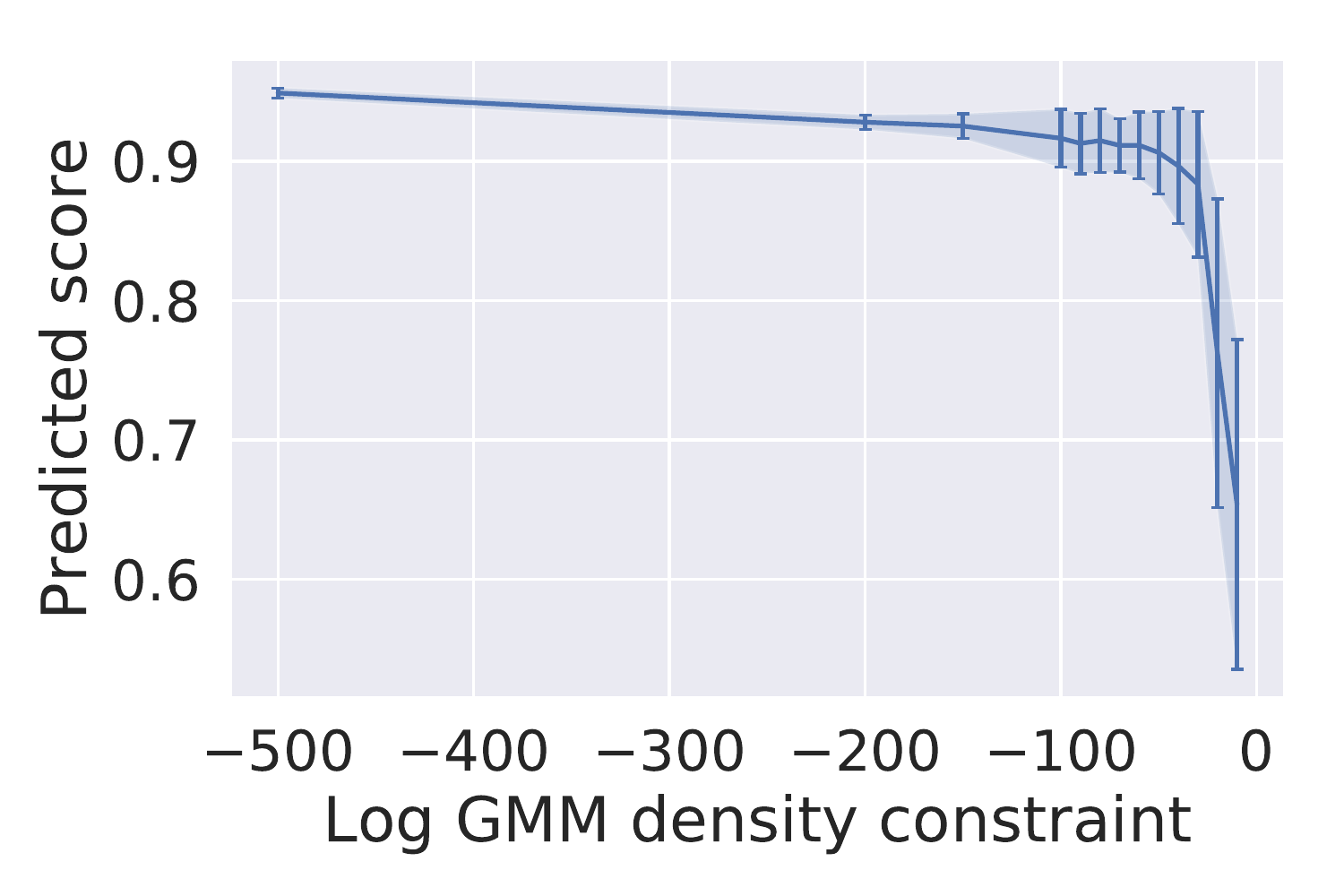}
        \caption{Thickness}
        \label{fig:predscores_thickness}
    \end{subfigure}
    \begin{subfigure}{0.33\textwidth}
        \includegraphics[width=\textwidth]{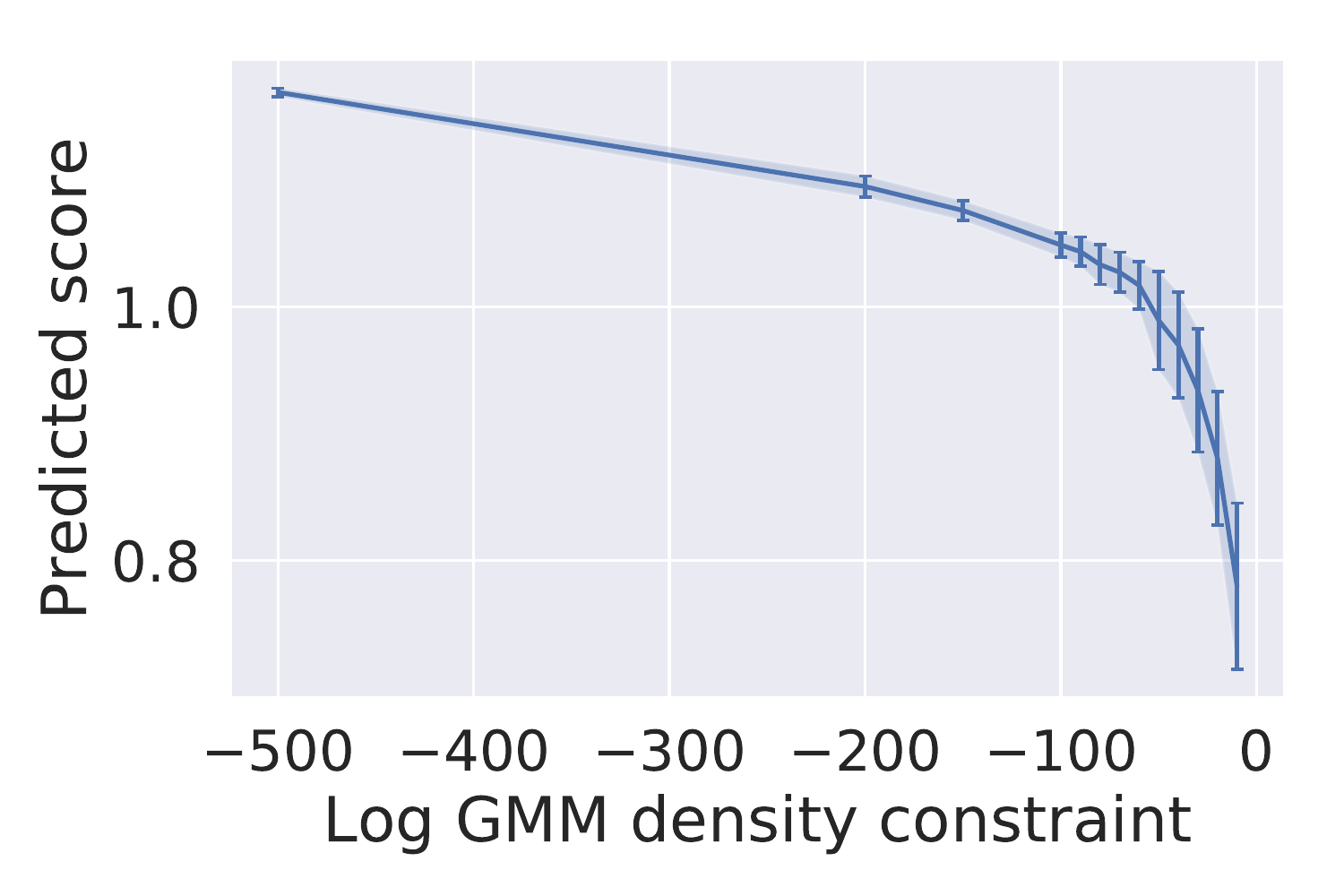}
        \caption{Aspect ratio}
        \label{fig:predscores_aspect}
    \end{subfigure}
    \caption{Predicted objective scores as a function of log GMM density constraint}
    \label{fig:predscores}
\end{figure}

\begin{figure}[h]
    \begin{subfigure}{0.33\textwidth}
        \includegraphics[width=\textwidth]{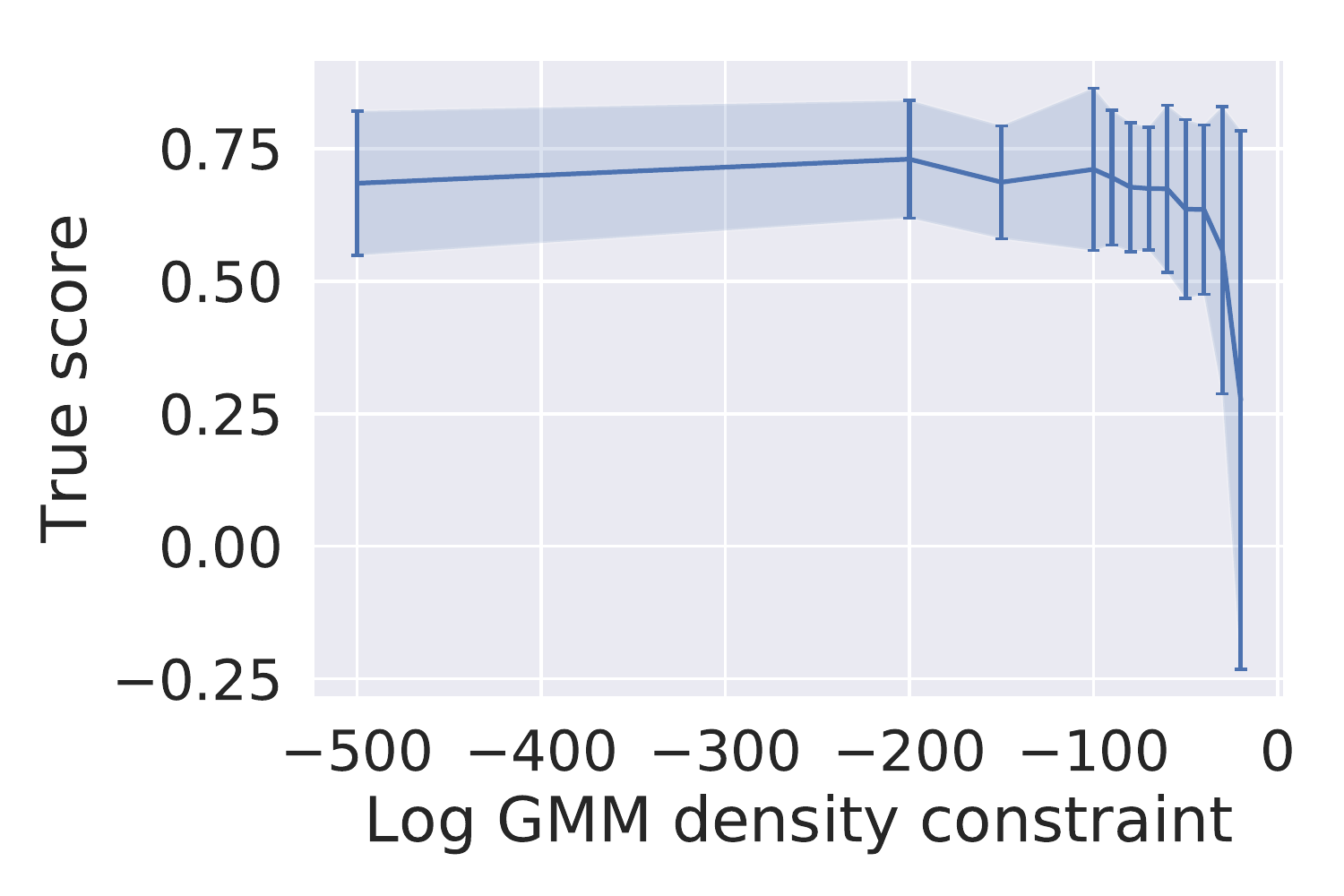}
        \caption{Rotation}
        \label{fig:truescores_pca}
    \end{subfigure}
    \begin{subfigure}{0.33\textwidth}
        \includegraphics[width=\textwidth]{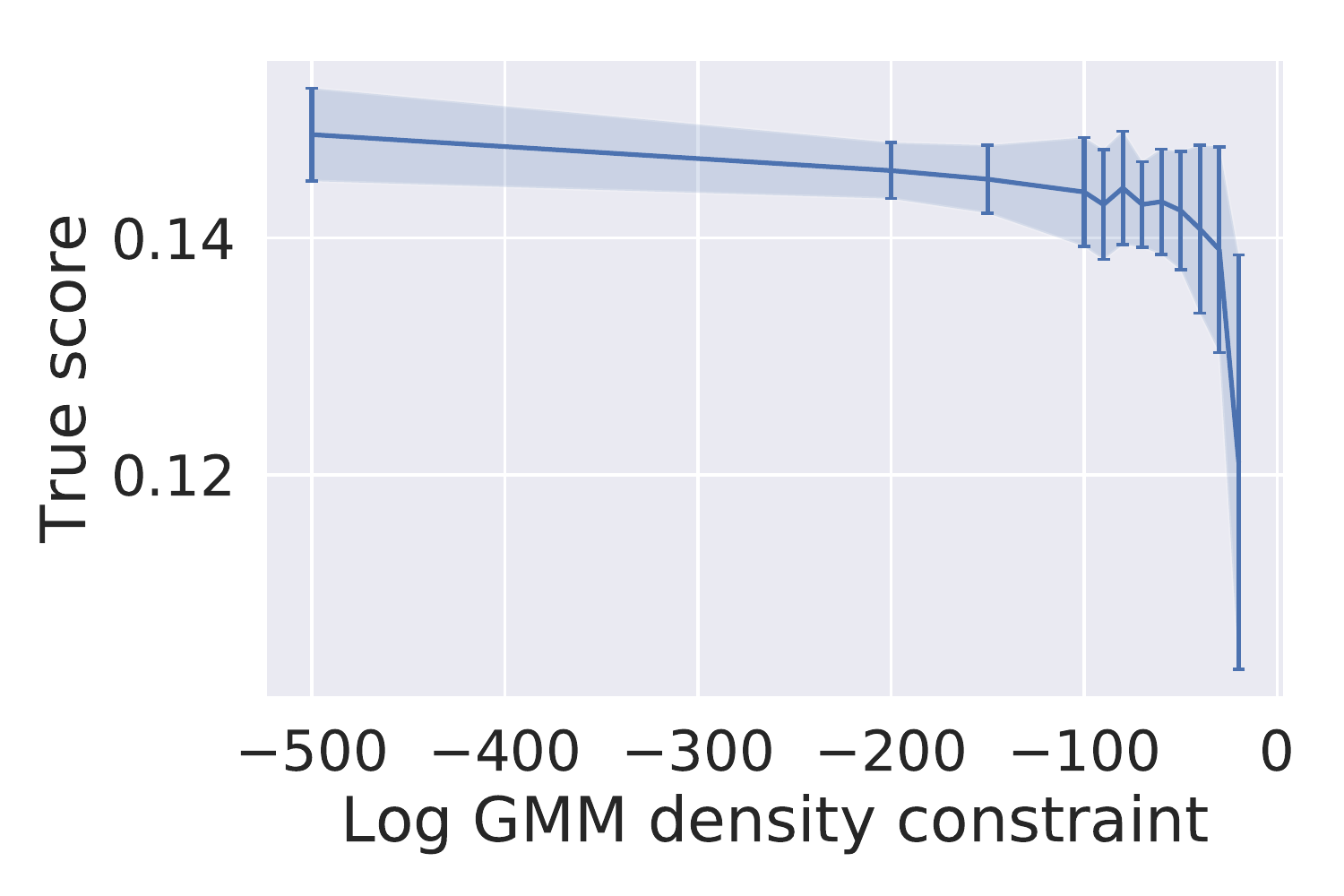}
        \caption{Thickness}
        \label{fig:truescores_thickness}
    \end{subfigure}
    \begin{subfigure}{0.33\textwidth}
        \includegraphics[width=\textwidth]{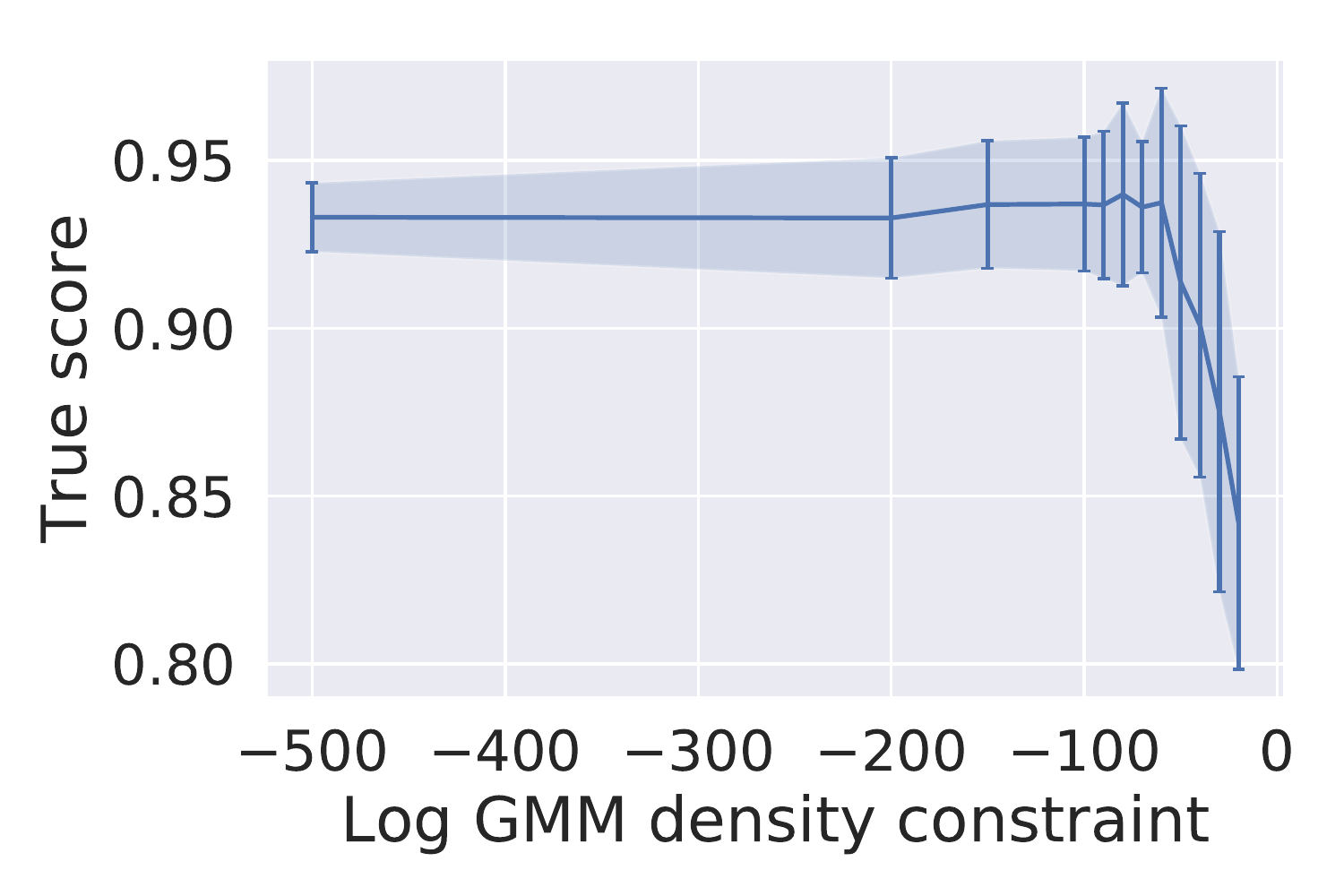}
        \caption{Aspect ratio}
        \label{fig:truescores_aspect}
    \end{subfigure}
    \caption{True objective scores as a function of log GMM density constraint}
    \label{fig:truescores}
\end{figure}

Figures \ref{fig:predscores} and \ref{fig:truescores} show the predicted and true scores of optimal points in the latent space respectively. The figures show that the predicted and true scores both decrease as the constraint is tightened.

\begin{figure}[!htbp]
    \centering
    \begin{subfigure}{0.24\textwidth}
        \centering
        \begin{subfigure}{0.45\textwidth}
            \includegraphics[width=\textwidth]{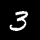}
            \label{fig:mnist_original_1}
        \end{subfigure}
        \begin{subfigure}{0.45\textwidth}
            \includegraphics[width=\textwidth]{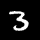}
            \label{fig:mnist_original_2}
        \end{subfigure}
        \caption{Training examples}
        \label{fig:mnist_training_examples}
    \end{subfigure}
    \begin{subfigure}{0.24\textwidth}
        \centering
        \begin{subfigure}{0.45\textwidth}
            \includegraphics[width=\textwidth]{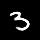}
            \label{fig:mnist_pca_-20}
        \end{subfigure}
        \begin{subfigure}{0.45\textwidth}
            \includegraphics[width=\textwidth]{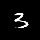}
            \label{fig:mnist_pca_-500}
        \end{subfigure}
        \caption{Rotation}
        \label{fig:mnist_pca_examples}
    \end{subfigure}
    \begin{subfigure}{0.24\textwidth}
        \centering
        \begin{subfigure}{0.45\textwidth}
            \includegraphics[width=\textwidth]{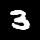}
            \label{fig:mnist_thickness_-20}
        \end{subfigure}
        \begin{subfigure}{0.45\textwidth}
            \includegraphics[width=\textwidth]{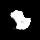}
            \label{fig:mnist_thickness_-500}
        \end{subfigure}
        \caption{Thickness}
        \label{fig:mnist_thickness_examples}
    \end{subfigure}
    \begin{subfigure}{0.24\textwidth}
        \centering
        \begin{subfigure}{0.45\textwidth}
            \includegraphics[width=\textwidth]{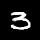}
            \label{fig:mnist_aspect_-20}
        \end{subfigure}
        \begin{subfigure}{0.45\textwidth}
            \includegraphics[width=\textwidth]{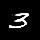}
            \label{fig:mnist_aspect_-500}
        \end{subfigure}
        \caption{Aspect ratio}
        \label{fig:mnist_aspect_examples}
    \end{subfigure}
    \caption{Training examples (\subref{fig:mnist_training_examples}) compared with samples from the optimisation process (\subref{fig:mnist_pca_examples}-\subref{fig:mnist_aspect_examples}). For (\subref{fig:mnist_pca_examples}-\subref{fig:mnist_aspect_examples}), the left and right figures correspond to $\eta=-20$ and $\eta=-500$ respectively.}
    \label{fig:mnist_examples}
\end{figure}

Figure \ref{fig:mnist_examples} shows examples of samples generated by COLD along with data points from the training set for illustration.

\begin{figure}[!htbp]
    \begin{subfigure}{0.33\textwidth}
        \includegraphics[width=\textwidth]{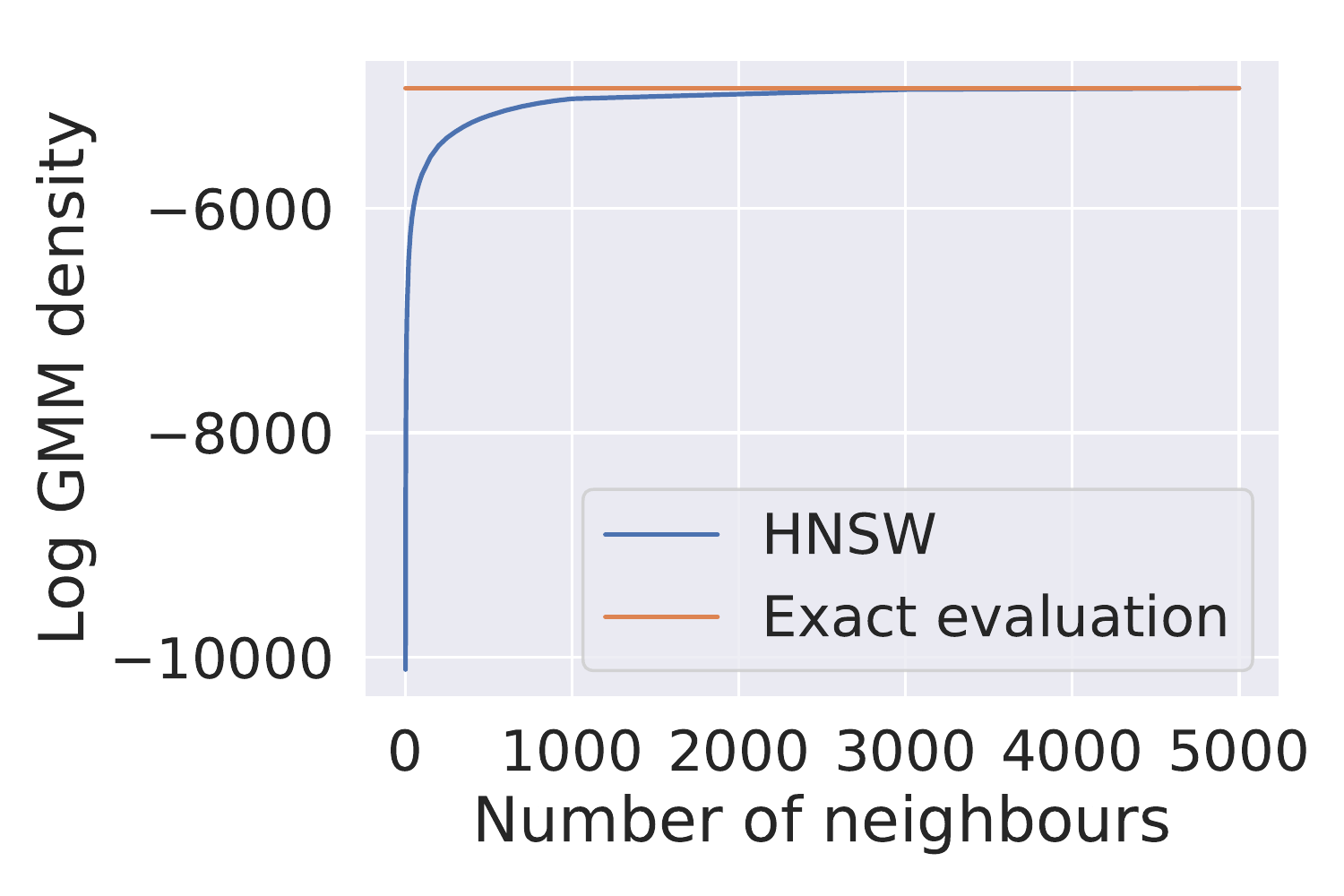}
        \caption{Rotation}
        \label{fig:knn_pca}
    \end{subfigure}
    \begin{subfigure}{0.33\textwidth}
        \includegraphics[width=\textwidth]{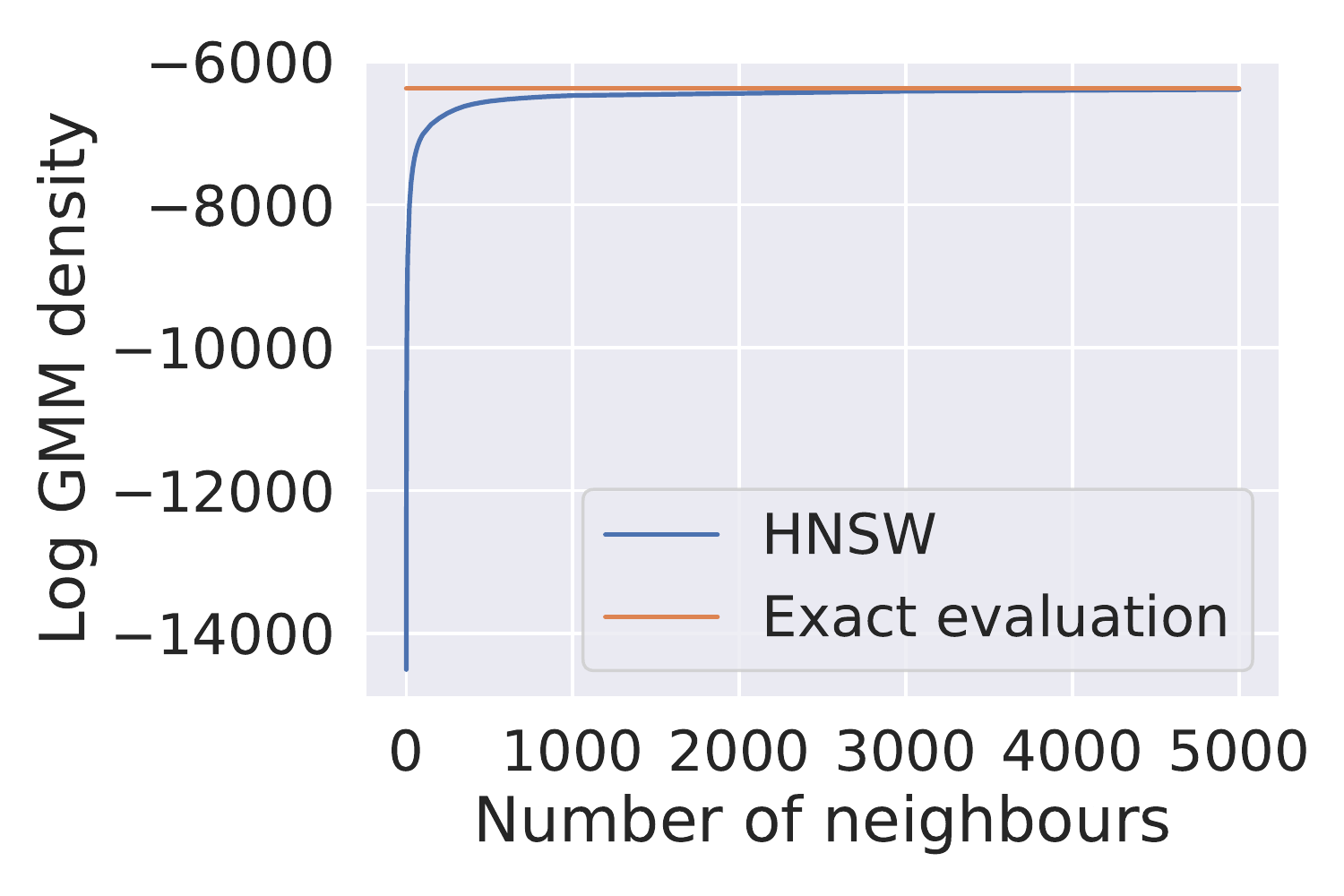}
        \caption{Thickness}
        \label{fig:knn_thickness}
    \end{subfigure}
    \begin{subfigure}{0.33\textwidth}
        \includegraphics[width=\textwidth]{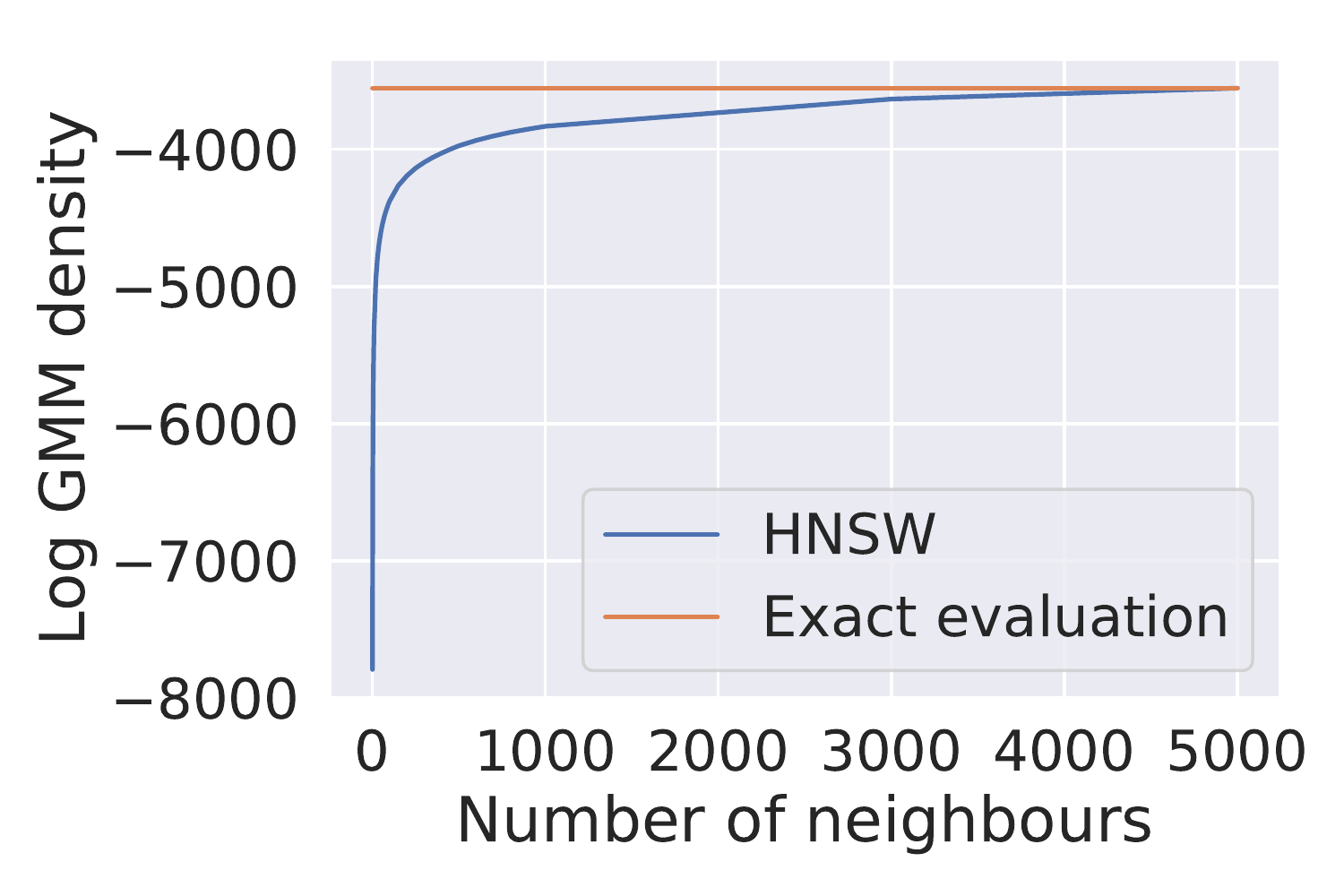}
        \caption{Aspect ratio}
        \label{fig:knn_aspect}
    \end{subfigure}
    \caption{KNN density as a function of number of neighbours, with true density for reference}
    \label{fig:knn}
\end{figure}

Finally, we measure the accuracy of the k-nearest neighbours approximation to the log GMM density. We calculate the mean density across all sampled grid points in the latent space for different values of $k$. The results are shown in Figure \ref{fig:knn}. In each plot, the horizontal orange line represents the mean of the true log GMM density, whereas the blue curve represents the mean of the approximate log GMM density. We observe that the approximate density approaches the true density as $k$ is increased.

\begin{figure}[!htbp]
    \begin{subfigure}{0.33\textwidth}
        \includegraphics[width=\textwidth]{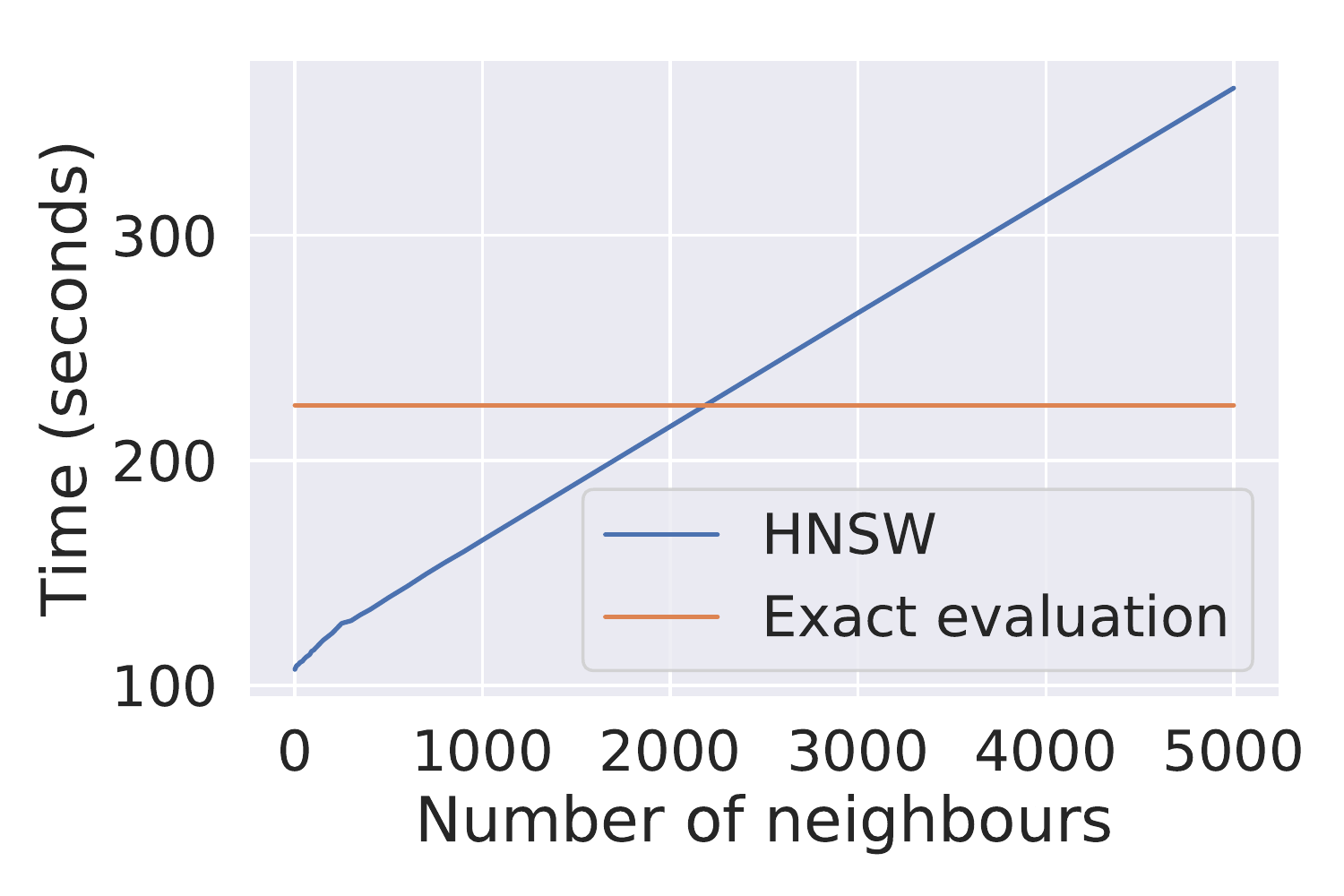}
        \caption{Rotation}
        \label{fig:knn_speed_pca}
    \end{subfigure}
    \begin{subfigure}{0.33\textwidth}
        \includegraphics[width=\textwidth]{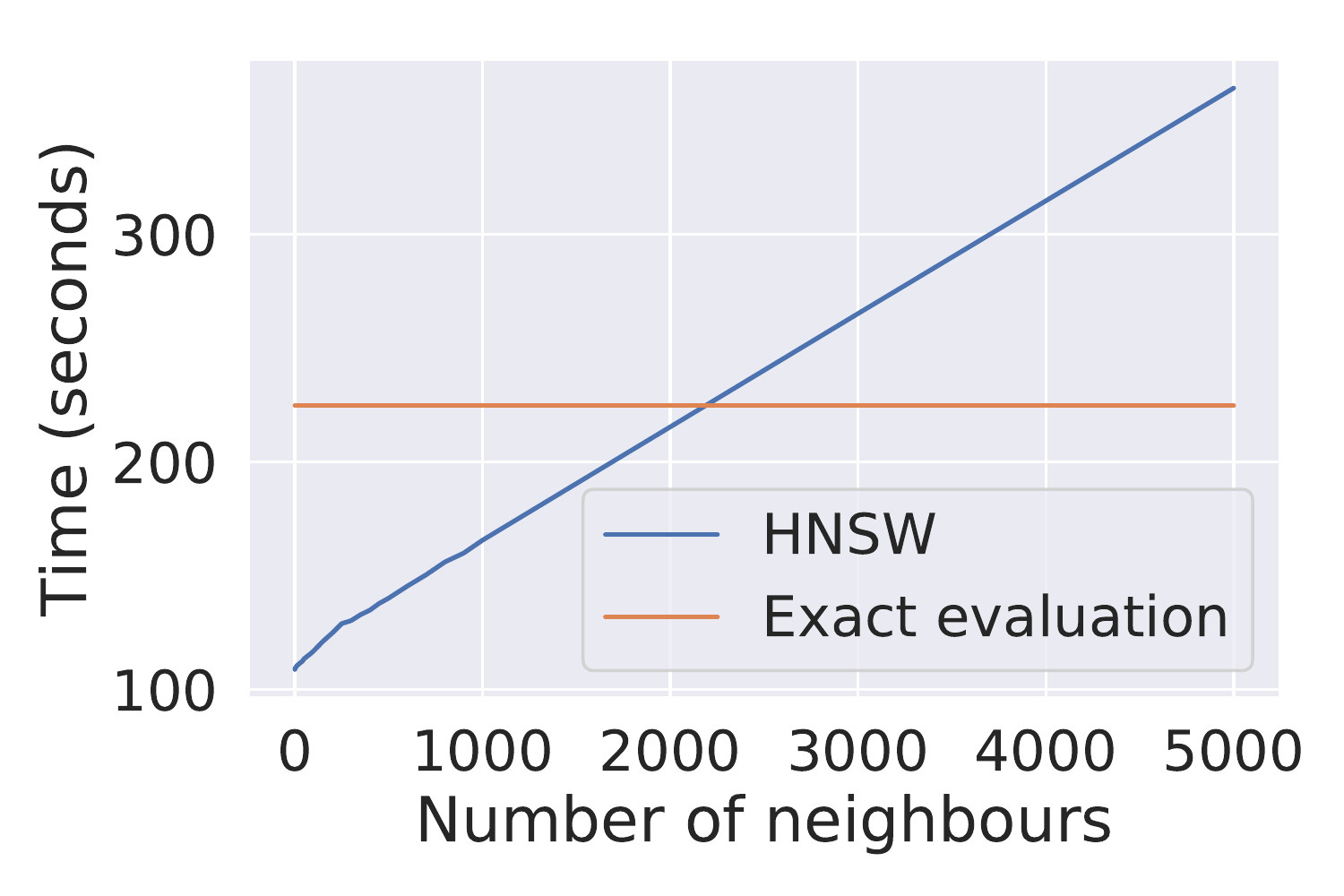}
        \caption{Thickness}
        \label{fig:knn_speed_thickness}
    \end{subfigure}
    \begin{subfigure}{0.33\textwidth}
        \includegraphics[width=\textwidth]{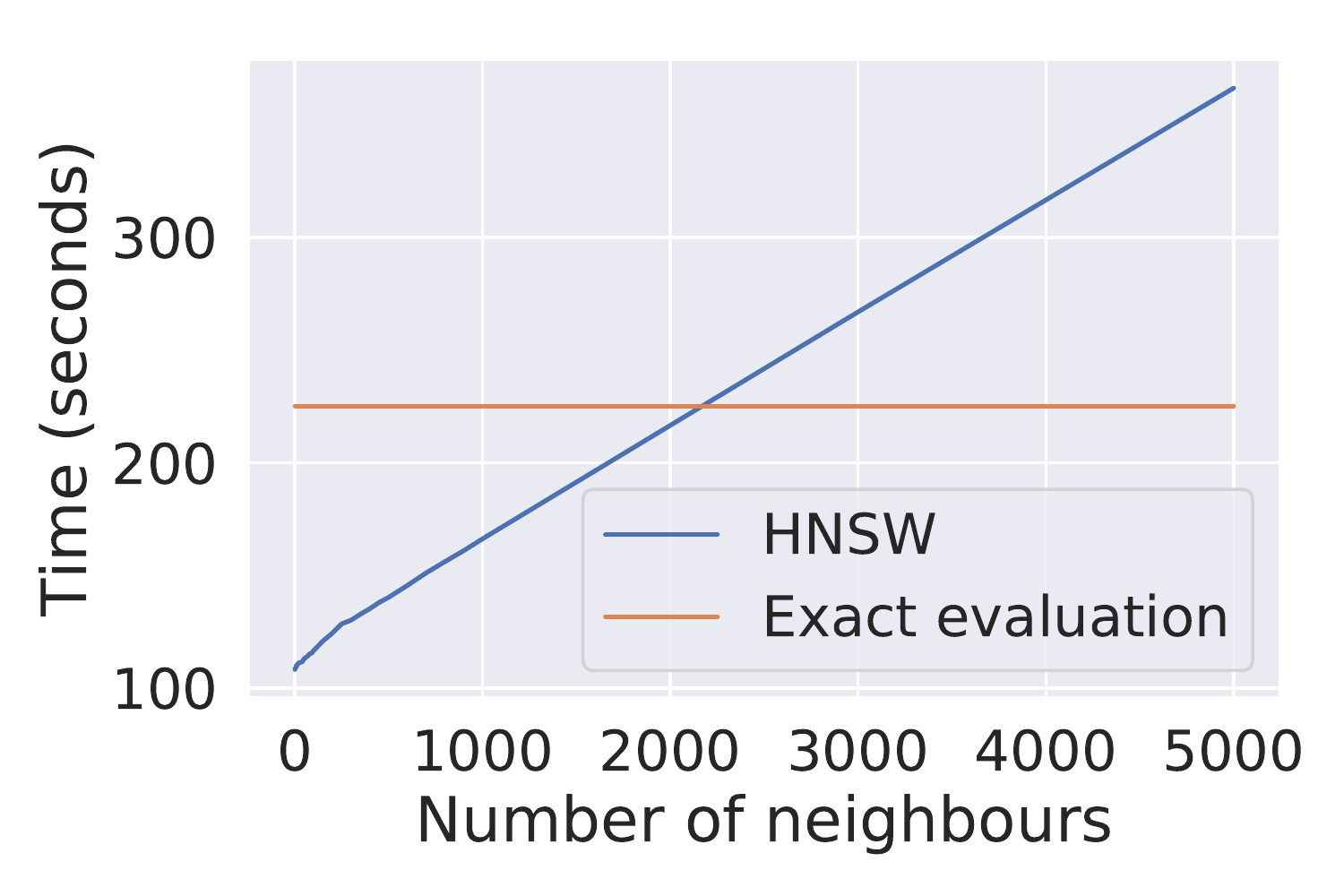}
        \caption{Aspect ratio}
        \label{fig:knn_speed_aspect}
    \end{subfigure}
    \caption{Time taken for KNN density calculation on 100,000 grid points as a function of number of neighbours. The time taken for exact evaluation is plotted as a horizontal line for reference.}
    \label{fig:knn_speed}
\end{figure}

We calculate the time taken to evaluate the density at 100,000 gridpoints using the HNSW method on a 2.40 GHz Intel Skylake CPU. For all three experiments, we also calculate the time taken to evaluate the exact density. The results are shown in Figure \ref{fig:knn_speed}. We find that the speed of the HNSW approach scales linearly with $k$. The HNSW approach is faster than exact evaluation for values of $k$ approximately less than 2000, and slower for greater values of $k$.

Our experiments show that the quality of the generated images increases as the log GMM density threshold is tightened, and that convergence does not occur in the absence of constraints. This shows that it is essential to enforce a constraint on the optimisation process.

The diversity of the images generally decreases as the constraint is loosened, which is likely due to the fact that looser constraints allow the optimiser to explore further away from the initialisation points in lower density areas, or cross low density valleys to reach other high density areas. This means that the dependence of the point of convergence on the point of initialisation is weakened, leading optimisation runs from different initial coordinates to converge to similar local optima. This hypothesis is supported by the fact that the predicted scores decrease as the constraint is tightened, which corresponds to less exploration of the latent space at tight constraints, including the inability to traverse low density areas.

Our experiments with the $k$ nearest neighbours approach to density estimation show that we can choose a value of $k < N$ experimentally in order to reduce computational complexity while preserving an acceptable level of accuracy. This method is likely to prove useful for values of $k$ that are not too large, as we see that exact evaluation is faster when $k$ gets large.

\subsection{Molecular Data}
\subsubsection{Experimental Setup}
We implement our method on molecules in the ChEMBL dataset\cite{chembl}. Our target property is the QED score, a value between 0 and 1 that expresses the `drug-likeness' of a molecule \cite{qed}.

Our PVAE is based on the work by Li et al \cite{graph_generative_model}, in which a conditional generative model for graphs is used to learn a probabilistic policy for sequential molecule generation. We use the components of this model to construct a VAE with a $N(z; 0, \mathbb{I})$ prior. (The code for these components is adapted from \cite{molecule_generator_repo}.) The dimensionality of $z$ is set to 100.

Each molecule in the training set is represented as the canonical series of operations used to construct it and passed through an encoder. The encoder consists of the graph convolutional network used by Li et al \cite{graph_generative_model}, followed by a fully connected network. The output from this network is the mean and diagonal covariance of a Gaussian encoding distribution $q(z|x_i)$. A sample $z$ is taken from the encoding distribution and used as the input to Li et al's conditional graph generative model \cite{graph_generative_model}. The decoder outputs the log of the likelihood $p(x_i|z)$, which is added to the KL term to yield the ELBO.

The predictor is a fully connected network with ReLU activation on the two hidden layers and sigmoid activation on the final layer. To ensure that the latent space structure accounts for our target property, we use a hyperparameter $\beta=0.00005$ to balance the gradients from the predictor and the ELBO:

\begin{equation}
Loss = -\beta \cdot ELBO + (1 - \beta) (w - f(z))^2
\end{equation}

After training the model, we calculate the approximate GMM density ($k=250$) at 1,000,000 points sampled from the prior distribution. We carry out optimisation using COBYLA \cite{powell_1994} with $t=10$ for different values of $\eta$ and grid points drawn from different distributions. The distributions used are similar to the prior; they are defined as $\mathcal{N}(0, b \cdot \mathbb{I})$, with $b \in \mathbb{R}$. We sample from these different distributions because lower variance distributions should yield points closer to the origin, around which the encoded training data is located due to the presence of the KL term in the ELBO during training. By changing $b$ we can change the distance of the sampled grid points from the training data, and hence obtain points with different GMM densities in different parts of the latent space.

\subsubsection{Results and Analysis}

\begin{figure}[!htbp]
\begin{subfigure}{0.5\textwidth}
\includegraphics[width=\textwidth]{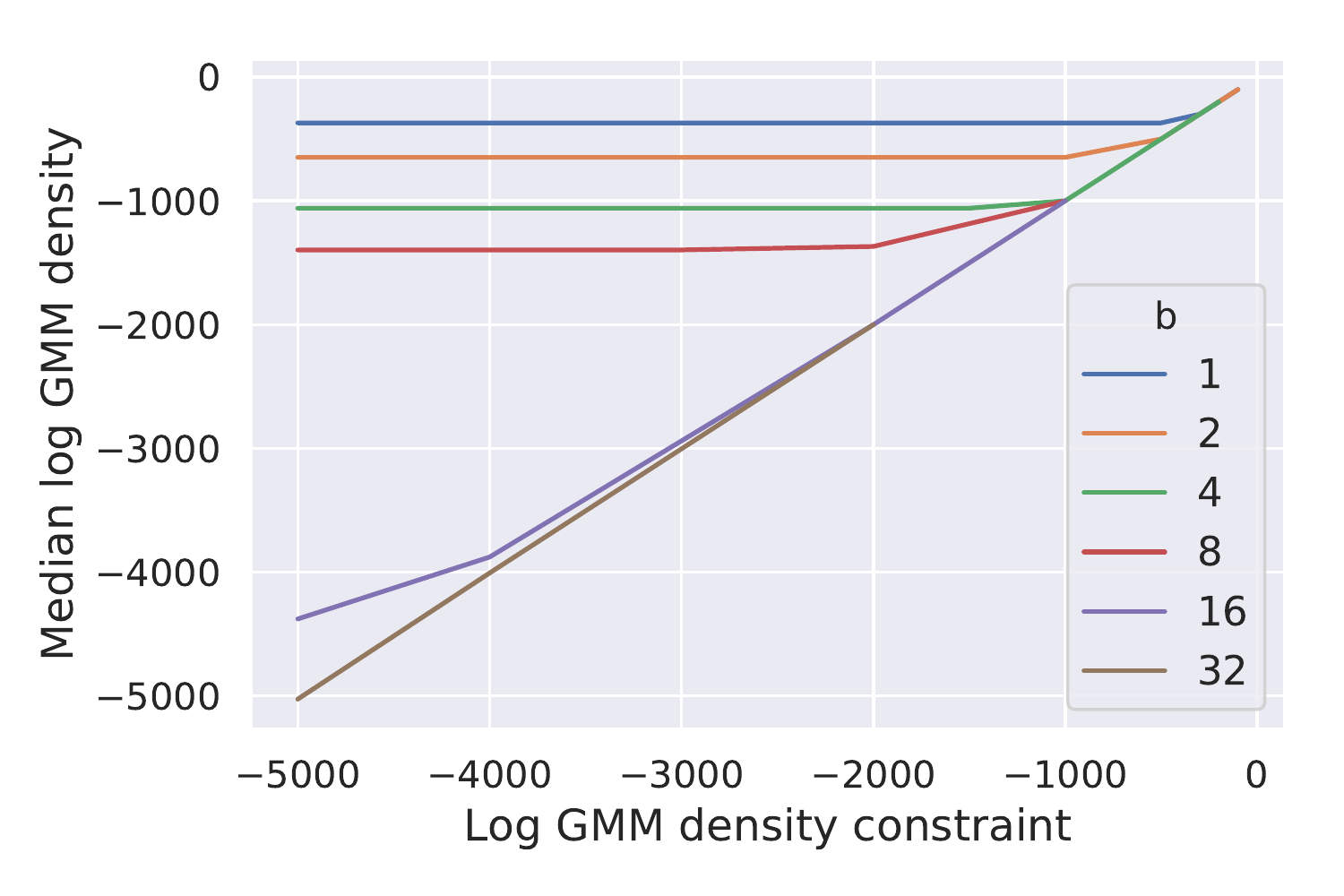}
\caption{Median log GMM density at optima}
\label{median_density}
\end{subfigure}
\begin{subfigure}{0.5\textwidth}
\includegraphics[width=\textwidth]{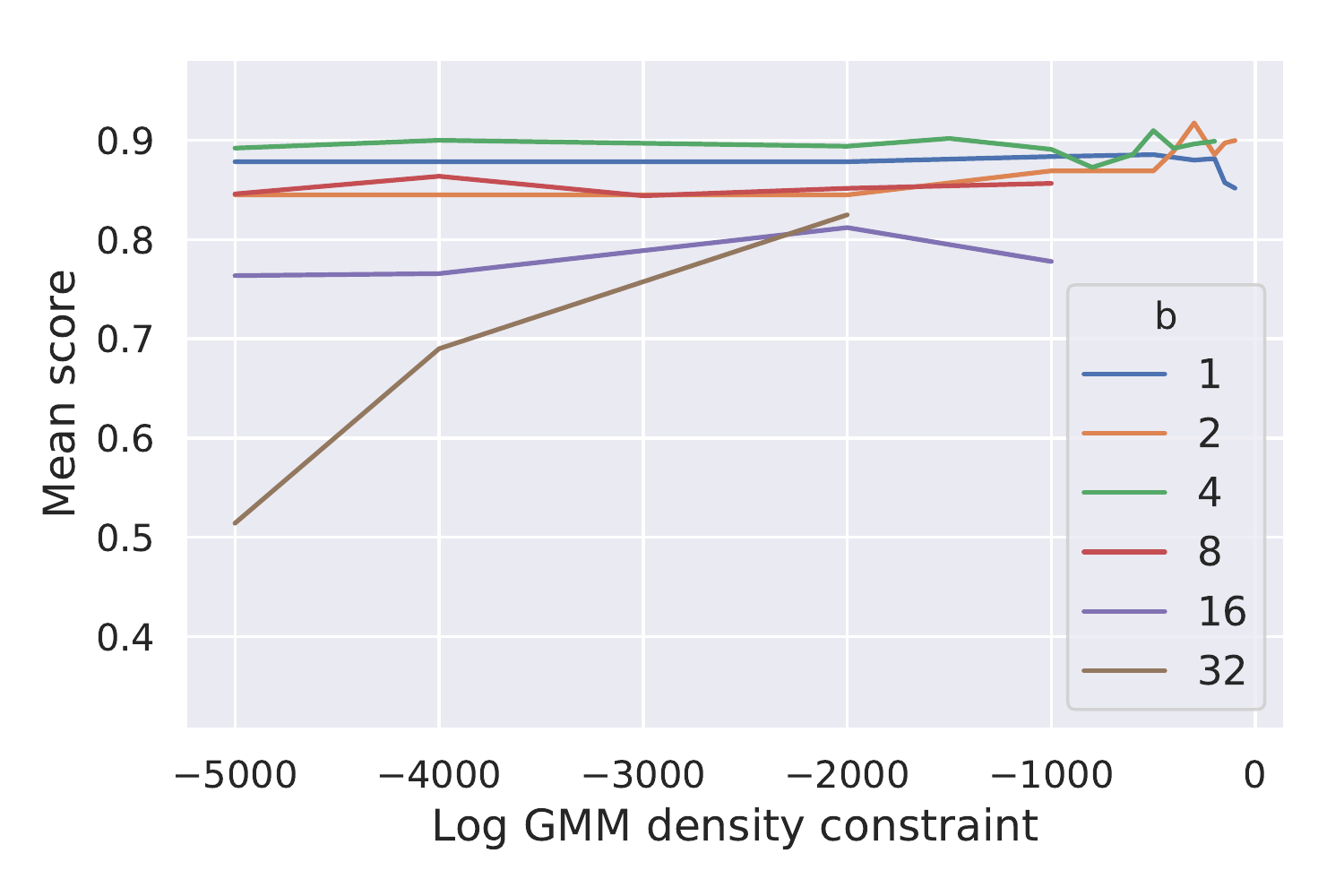}
\caption{Mean true predictor score at optima}
\label{mean_score}
\end{subfigure}
\begin{subfigure}{0.5\textwidth}
\includegraphics[width=\textwidth]{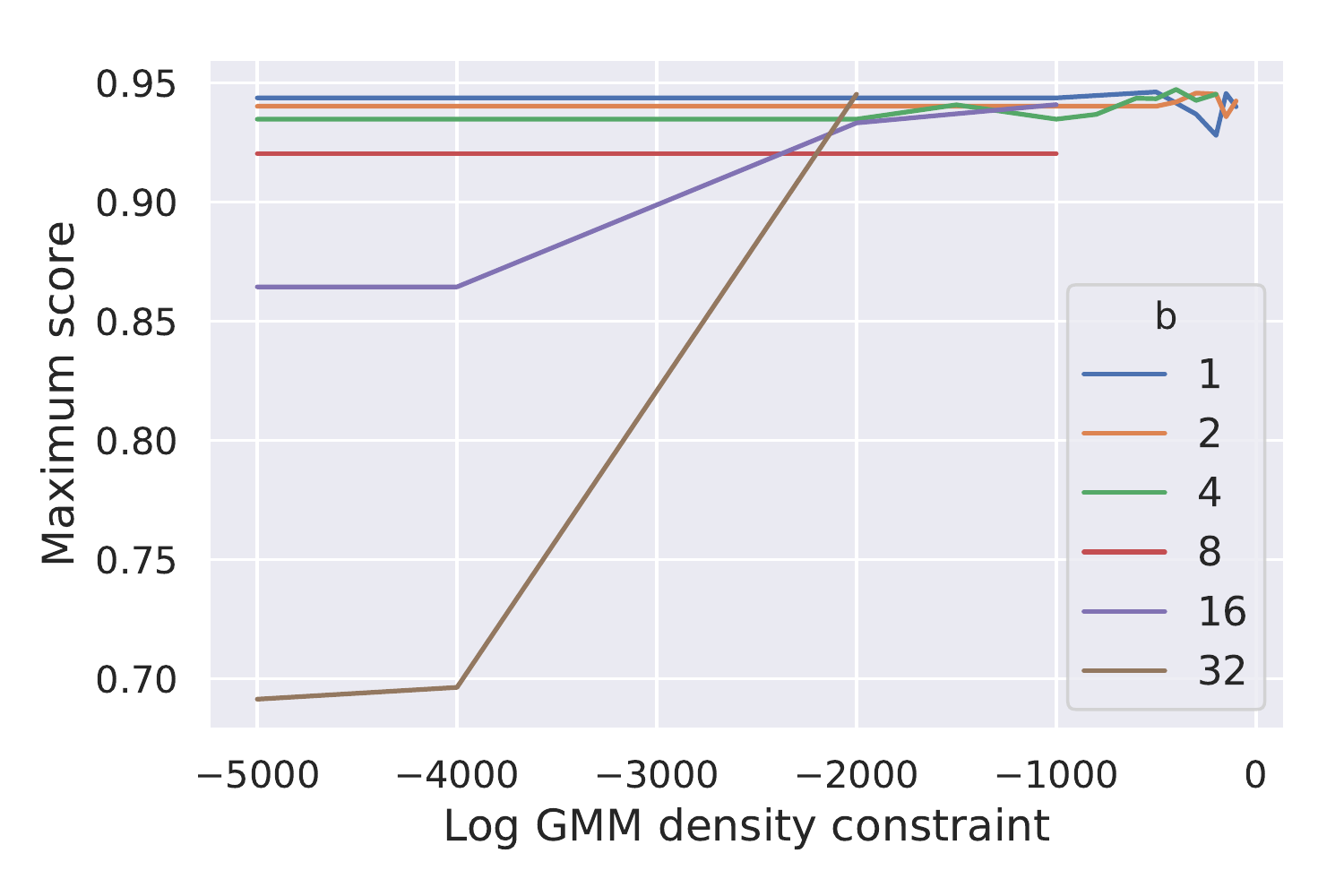}
\caption{Maximum of true scores at optima}
\label{maximum_score}
\end{subfigure}
\begin{subfigure}{0.5\textwidth}
\includegraphics[width=\textwidth]{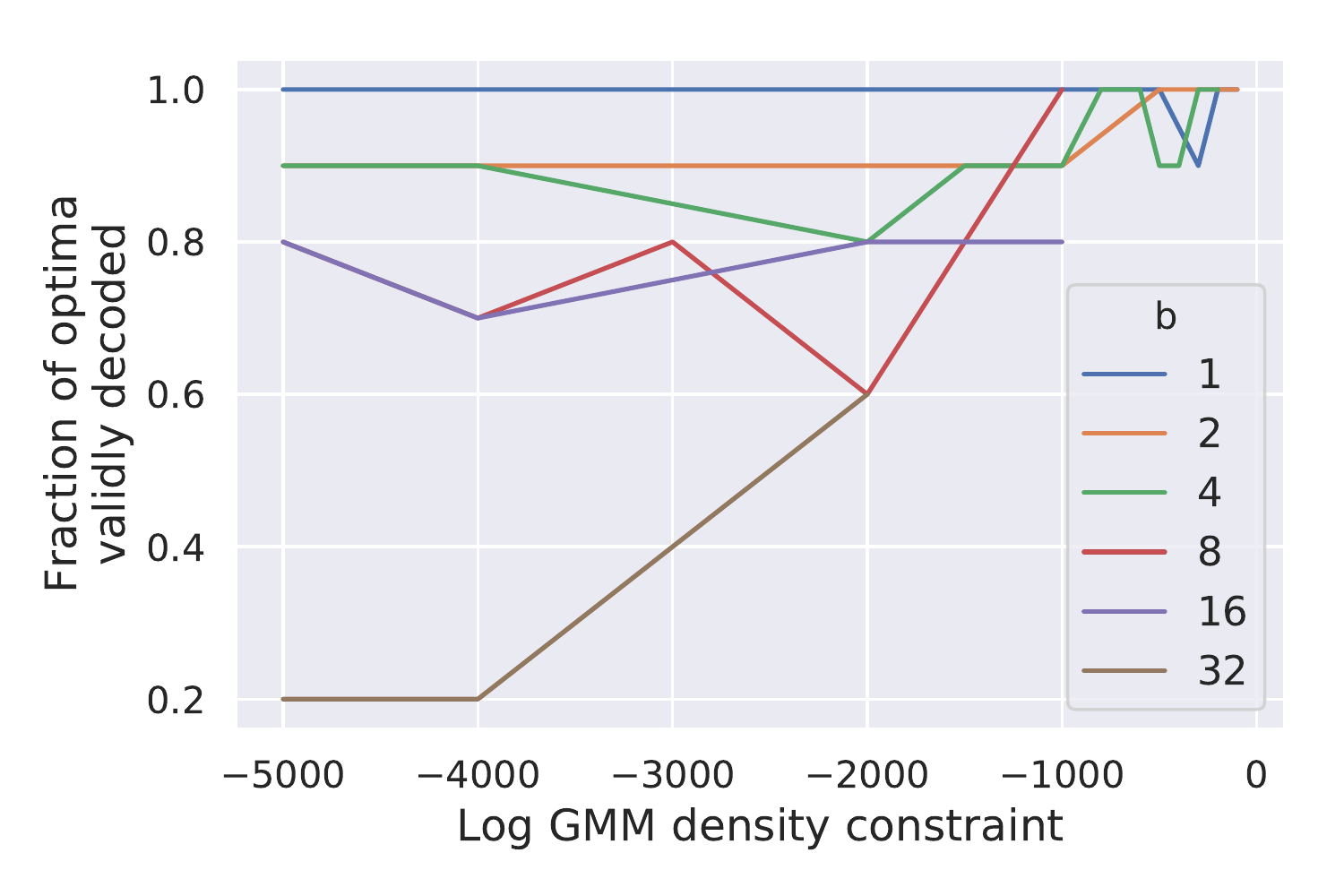}
\caption{Fraction of optima decoded to valid molecules}
\label{frac_valid_optima}
\end{subfigure}
\caption{Statistics for ChEMBL results with gridpoints generated using different values of $b$}
\label{optimisation_results}
\end{figure}

Figure \ref{optimisation_results} shows the results of the optimisation process. Validly decoded latent space points yield molecules when passed through the decoder and do not raise an exception. Figure \ref{median_density} shows that in the range of log densities considered, the median log density of the 10 optima for $b \leq 8$ eventually plateaus as $\eta$ decreases. The value of $\eta$ at which the density plateaus decreases as $b$ increases. Hence as the (pre-optimisation) log density of the grid points decreases, the value of $\eta$ at which the log density plateaus decreases. This shows that if local optimisation starts in a high density region, the local optimum found is more likely to be in a relatively high density region than if local optimisation starts in a low density region.

This finding is further supported by Figure \ref{mean_score}, which shows that in the majority of cases the mean score of the decoded optimum molecules plateaus beyond certain values of $\eta$. This is because as $\eta$ decreases, the constraint becomes inactive for an increasing number of the 10 pre-local optimisation optima as indicated by the discussion above. Once the constraint is inactive for a particular starting point, the starting point will converge to the same optimum even if the threshold is lowered. Decreasing the threshold further will only change the mean score if this starting point is displaced from the top 10 pre-local optimisation optima by a starting point with a higher score but lower density. This also explains why the mean score plateaus at decreasing values as $b$ and hence the mean initial log density decreases; there is a greater proportion of starting points with lower densities in the samples corresponding to high $b$ than there is in those corresponding to low $b$. Hence using the constraint in the optimisation process is more useful when sampling points over a diverse area than when sampling points in a region where most of the encoded means are relatively densely located.

Figure \ref{maximum_score} shows that the true maximum score at the optima also plateaus beyond certain values of $\eta$ as the same molecule is obtained at the optimum. Figure \ref{frac_valid_optima} shows that the fraction of successfully decoded molecules generally decreases as $\eta$ decreases (with the exception of the case where $b = 1$ and all sampled points are in relatively high density regions).

\begin{figure}[!h]
\includegraphics[width=\textwidth]{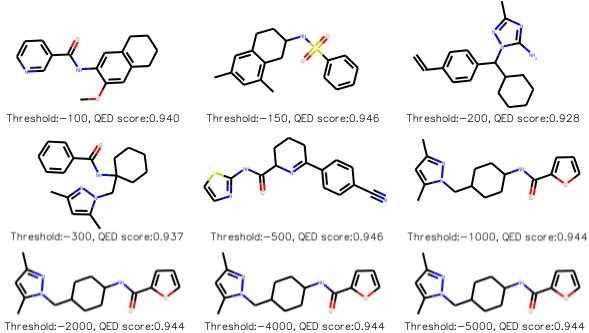}
\caption{Molecules with highest QED scores for different threshold values for grid points with $b=1$}
\label{var_1_best}
\end{figure}

We hence observe the convergence of several optimisations with the same starting constraint to the same optimum even in the presence of very loose optimisation constraints. This may be because the surface of the predictor function in the latent space is highly curved, implying sharp local optima. Using a different predictor architecture with fewer parameters may mitigate this problem and cause the constraint to play a more significant part in local optimisation.

The best molecules obtained for each threshold value for the grid points corresponding to $b = 1$ are shown in Figure \ref{var_1_best} along with their QED scores. We can see the best molecule change as the threshold decreases from -100 to -1000, but from a threshold of -1000 to -5000 the best molecule remains the same, confirming our discussion above. The QED scores of even the molecules with low thresholds are high, confirming that with this latent space structure, if optimisation starts in a high density area then high-scoring molecules may be produced regardless of the local optimisation constraint.

\begin{figure}[!h]
\includegraphics[width=\textwidth]{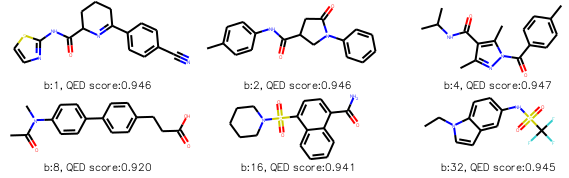}
\caption{Molecules with highest QED scores for grid points with different values of $b$}
\label{each_var_best}
\end{figure}

The best molecules obtained from gridpoints corresponding to different values of $b$ are shown in Figure \ref{each_var_best}. We observe differences in molecular structure as we change $b$. This confirms that different parts of the latent space can yield molecules that are structurally different but which have similarly high scores. The highest QED score for a molecule in the training data is 0.9483, so our model can produce molecules that have scores similar to the highest scores for molecules in the training data. A comparison of SMILES strings shows that none of the molecules generated by COLD are present in the training data. COLD can therefore be used to create novel high-scoring molecules.

\vspace{-\baselineskip}
\section{Conclusion and Future Work}
We have shown that a Gaussian mixture model can be effectively used to approximate how well different parts of the latent space have been trained. The density of a GMM composed of the encoding distributions of the training data can be used to filter initial latent space points that can be used reliably for local optimisation. Constrained local optimisation, using the constraint that the GMM density be above a certain value, can be used to optimise in the space around high-scoring points to yield higher-scoring points that correspond to novel high-scoring molecules. The HSNW method for k-nearest neighbours approximation can be used to make the process more computationally efficient.

A main focus of future work could be to find a better way of sampling points from the latent space. A limitation of the process here is that an entire grid cannot be loaded into memory, so grid points are sampled from the latent space. This yields grid points that may not necessarily be good starting points for local optimisation. The choice of the variance of the sampling distribution also strongly determines the range of densities of the sampled points for any reasonable number of sampled points, with lower variances yielding higher density points. Sampling in a way that yields points at a large range of densities, both low and high, is an essential next step in carrying out truly global optimisation.

\bibliographystyle{plain}
\bibliography{neurips_2019}

\end{document}